\documentclass[10pt,journal,compsoc]{IEEEtran}

\ifCLASSOPTIONcompsoc
  \usepackage[nocompress]{cite}
\else
  \usepackage{cite}
\fi

\usepackage{url}
\usepackage{latexsym}
\usepackage{booktabs}
\usepackage{paralist}
\usepackage{graphicx}
\usepackage{subcaption}
\usepackage[table]{xcolor}
\usepackage{fancyhdr}

\usepackage{color}

\fancypagestyle{empty}{%
  \fancyhf{}
  \fancyhead[L]{\small {\color{blue} This is the accepted version of an article accepted for publication in \textit{IEEE Intelligent Systems @ 2018 IEEE}}}
}

\begin{document}
%
\title{Political Homophily in Independence Movements: Analysing and Classifying Social Media Users by National Identity}
%
%
%
%

\author{Arkaitz~Zubiaga,
        Bo~Wang,
        Maria~Liakata,
        and~Rob~Procter
\IEEEcompsocitemizethanks{\IEEEcompsocthanksitem A. Zubiaga, B. Wang, M. Liakata and R. Procter are with the Department of Computer Science, University of Warwick, Gibbet Hill Road, Coventry CV4 7AL, United Kingdom. B. Wang, M. Liakata and R. Procter are also with the Alan Turing Institute, 96 Euston Rd, Kings Cross, London NW1 2DB, United Kingdom.\protect\\
E-mail: see http://www.zubiaga.org/
}
}

\markboth{}
{Zubiaga \MakeLowercase{\textit{et al.}}: Political Homophily in Independence Movements: Analysing and Classifying Social Media Users by National Identity}



\IEEEtitleabstractindextext{%
\begin{abstract}
 Social media and data mining are increasingly being used to analyse political and societal issues. Here we undertake the classification of social media users as supporting or opposing ongoing independence movements in their territories. Independence movements occur in territories whose citizens have conflicting national identities; users with opposing national identities will then support or oppose the sense of being part of an independent nation that differs from the officially recognised country. We describe a methodology that relies on users' self-reported location to build large-scale datasets for three territories -- Catalonia, the Basque Country and Scotland. An analysis of these datasets shows that homophily plays an important role in determining who people connect with, as users predominantly choose to follow and interact with others from the same national identity. We show that a classifier relying on users' follow networks can achieve accurate, language-independent classification performances ranging from 85\% to 97\% for the three territories.
\end{abstract}

}

\maketitle

\IEEEdisplaynontitleabstractindextext

%
\IEEEpeerreviewmaketitle

\thispagestyle{empty}


\IEEEraisesectionheading{\section{Introduction}\label{sec:introduction}}

%
%
%
%

\IEEEPARstart{S}{ocial} media are an increasingly important source for data mining applications, among others for exploratory research utilised as a means to analyse political and societal issues. One problem with social media is the limited availability of users' socio-demographic details that would enable analysis of the many different realities in society. Attempting to mitigate this issue, a growing body of research deals with the automated inference of socio-demographic characteristics such as age and gender \cite{rao2010classifying}, country of origin \cite{zubiaga2017towards} or political orientation \cite{pennacchiotti2011democrats}.

Following this line of research, we describe and assess a data collection methodology that enables identifying two groups of social media users in territories with active independence movements: those who support the independence (pro-independence), and those who oppose it (anti-independence). Independence movements are motivated by conflicting national identities, where different parts of a population identify themselves as citizens of one nation or another, such as the Scots feeling Scottish (pro-independence) or British (anti-independence). These situations lead to people with conflicting national identities living together in the same territory, where national identity can be defined as ``a body of people who feel that they are a nation'' \cite{emerson1962empire}.

Our study makes the following novel contributions: (1) we describe a methodology that relies on Twitter users' self-reported location for collecting users with conflicting national identities, as opposed to the largely studied partisanship or voting intention of users, (2) we perform a quantitative analysis focusing on the network and interactions within and across national identities, and (3) we study language-independent classification approaches using four different types of features. Our semi-automated data collection and annotation methodology enables us to collect datasets for three territories --Catalonia, the Basque Country and Scotland-- with over 36,000 users. Our experiments show that the users' network can achieve highly accurate classification, outperforming the use of tweet content. An analysis of the user groups highlights the influence of political homophily in independence movements, where users predominantly form ties on the basis of their ideology, following and interacting with others that think alike.

\section{Related Work}
\label{sec:related-work}

Computational approaches to the study of independence movements are scarce. The most relevant work to that which we report here is by Fang et al. \cite{fang2015topic} attempting to classify users' voting intention in the 2014 Scottish independence referendum. However, their work focused on determining voting intention during a particular referendum rather than determining the users' national identity and, being limited to a single territory -- Scotland --, they introduced a language-dependent approach that identifies topics discussed during the referendum campaign for determining users' stance. While classification of users by political orientation is also related to our work, such as republicans or democrats in the US \cite{pennacchiotti2011democrats}, or conservatives or labourists in the UK \cite{boutet2012s}, national identities reflect independent dimensions that are not necessarily linked to partisanship. Citizens with common national identities can also vote for parties with different political ideologies, and their national identities can be instead motivated by cultural and linguistic backgrounds \cite{duany2000nation}. Similarly, there has been research in predicting the outcome of political elections \cite{tsakalidis2015predicting,tumasjan2011election}, but this line of research again looks at the voting intention of users rather than their national identity.

Previous research has suggested that political homophily is also reflected in social media \cite{himelboim2013birds,barbera2015birds}, that is that supporters of one political party are more likely to follow one another than to follow supporters of other parties. Whether this generalises to users with different national identities has not been explored before.

\section{Data Collection}
\label{sec:data-collection}

Our data collection methodology relies on users' self-reported location as a proxy for identifying the territory that users claim to be citizens of, which is directly indicative of their stance towards the ongoing independence movement in their territory. For each territory, we identify distinctive location names with which either pro-independence or anti-independence people associate themselves, which gives us ground truth labels:

\noindent \textbf{Catalonia.} Citizens of Catalonia can feel either Catalan (pro-independence) or Spanish (anti-independence). For the generation of the dataset distinguishing these two national identities, we rely on the fact that Catalans whose profile location contains \textit{Pa\"isos Catalans} or its acronym \textit{PPCC} (i.e. Catalan Countries) are overtly claiming to be citizens of an independent Catalonia. The term \textit{Pa\"isos Catalans} unambiguously refers to an independent Catalonia, which would instead be \textit{Catalunya} or \textit{Catalu\~na} if not explicitly referring to an independent country. Alternatively, we identify users whose location contains the name of a Catalan city (e.g. Barcelona or Girona) or Catalunya/Catalu\~na along with \textit{Espanya} or \textit{Espa\~na} as claiming to be Spanish citizens. Using a dataset of 12 months' worth of tweets collected from the Twitter streaming API between March 2015 and February 2016, we sampled users that satisfied the above characteristics.

\noindent \textbf{Basque Country.} Citizens of the Basque Country can feel either Basque (pro-independence) or Spanish (anti-independence). To generate the dataset, we look for users whose profile location contains \textit{Euskal Herria} or its acronym \textit{EH} (i.e. Greater Basque Country). The term \textit{Euskal Herria} unambiguously refers to an independent Basque Country, unlike \textit{Euskadi} which refers to a region of Spain. On the other hand, we look for users whose location field contains the name of a Basque city (e.g. Bilbao or Donostia/San Sebasti\'an) or Euskadi along with \textit{Espainia} or \textit{Espa\~na}, which identifies users located in the Basque Country who claim to be citizens of Spain. We use the same 12 month dataset to look for users that satisfy these characteristics.

\noindent \textbf{Scotland.} Officially part of the UK, Scotland also has an ongoing independence movement. The dataset generation process for Scotland needs to be slightly different from the two above, as the Scots do not use a different name to refer to an independent Scotland. To overcome this, we first use a Twitter dataset pertaining to the 2014 Scottish independence referendum, collected between 1st August and 30th September, 2014 using a list of keywords including `\textit{\#IndyRef}', `\textit{vote}' and `\textit{referendum}'. In this dataset, we look for supporters who tweeted one of \textit{\#YesBecause, \#YesScotland, \#YesScot, \#VoteYes} and opposers who tweeted one of \textit{\#NoBecause, \#BetterTogether, \#VoteNo, \#NoThanks}, as suggested by \cite{fang2015topic}. To make sure that we identify the users' stance towards Scotland's independence, avoiding noise from tweets that are not necessarily endorsements of the hashtag being used, we collected the profile metadata of all sampled users. To generate the final dataset, we used again the same 12 month dataset, from which we retained the profiles of all IndyRef supporters whose profile location contained Scotland but not UK, United Kingdom, GB or Great Britain, as well as all opposers whose profile location contained the name of a Scottish city (e.g. Glasgow or Edinburgh) or Scotland, along with UK, United Kingdom, GB or Great Britain.

The location strings for the resulting user profiles were manually verified. The methodology was largely accurate, with 96.0\%, 95.9\% and 98.9\% correct instances for Catalonia, the Basque Country and Scotland, respectively. Those users that did not meet our expected locations were manually removed from the datasets. The resulting datasets consist of 36,609 users (see Table \ref{tab:dataset}).

\begin{table}[htb]
 \begin{center}
  \begin{tabular}{l l l l}
   \toprule
   & Pro-Independence & Anti-Independence & Total \\
   \midrule
   Catalonia & 2,361 & 8,599 & 10,960 \\
   \midrule
   Basque Country & 5,377 & 2,033 & 7,410 \\
   \midrule
   Scotland & 13,114 & 5,125 & 18,239 \\
   \bottomrule
   TOTAL & 20,852 & 15,757 & 36,609 \\
   \bottomrule
  \end{tabular}
  \caption{Distribution of users identified as pro-independence or anti-independence.}
  \label{tab:dataset}
 \end{center}
\end{table}

\subsection{User Data Collection}
\label{ssec:user-data}

For each user in our dataset, we collect three different types of data: (1) the user's 500 most recent tweets, (2) the 500 most recent tweets favourited by the user, and (3) the list of users that the user follows and is followed by. The final collection comprises 27.4 million tweets including timelines and favourites, as well as 19.1 million different users occurring in follow networks.

\section{Analysis of National Identity Groups}
\label{sec:analysis}

\begin{figure*}[tbh]
 \centering
 \begin{subfigure}[b]{0.33\textwidth}
  \includegraphics[width=\textwidth]{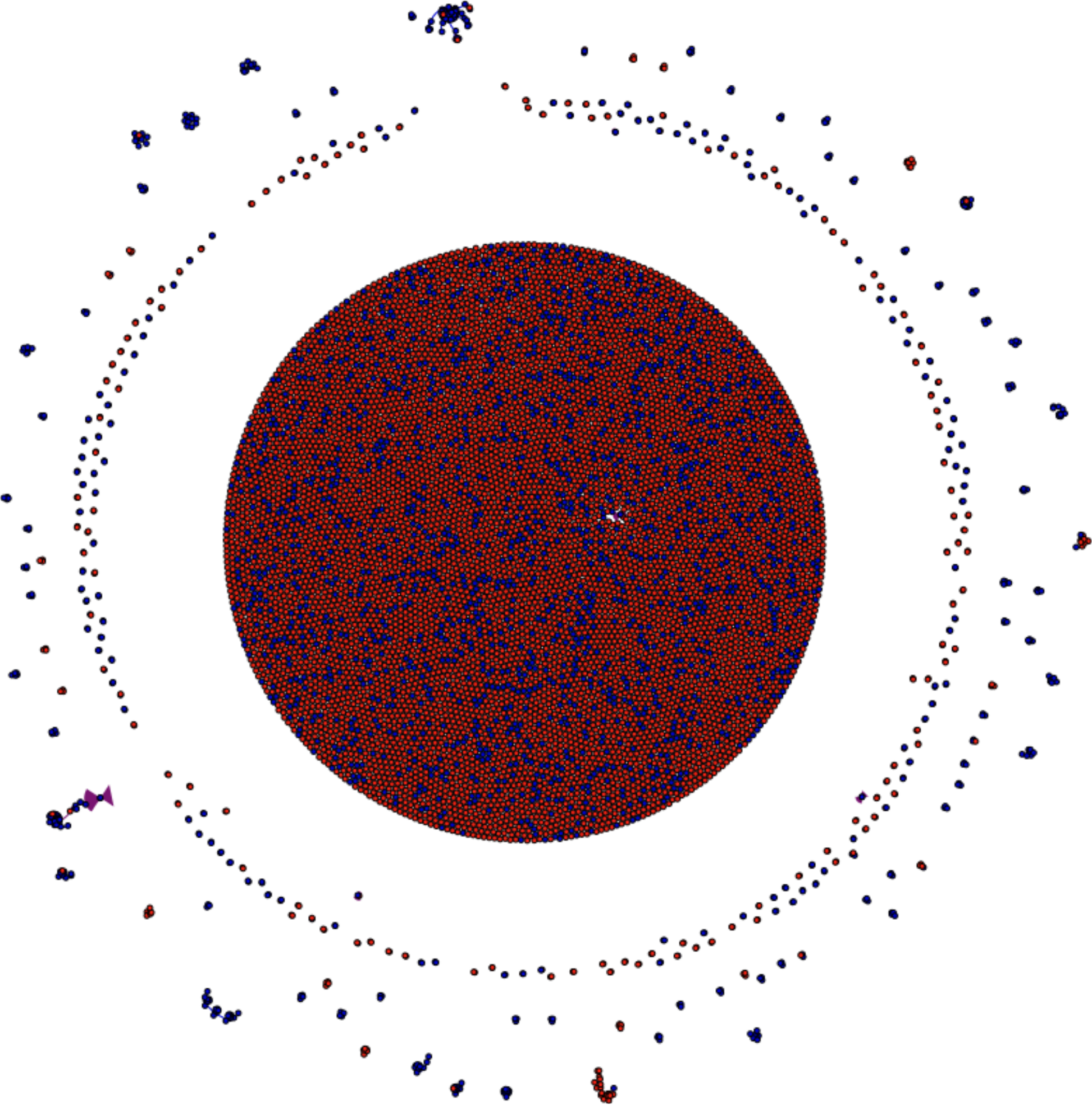}
  \caption{Interactions (Catalonia)}
  \label{fig:interactions-ca}
 \end{subfigure}
 \begin{subfigure}[b]{0.33\textwidth}
  \includegraphics[width=\textwidth]{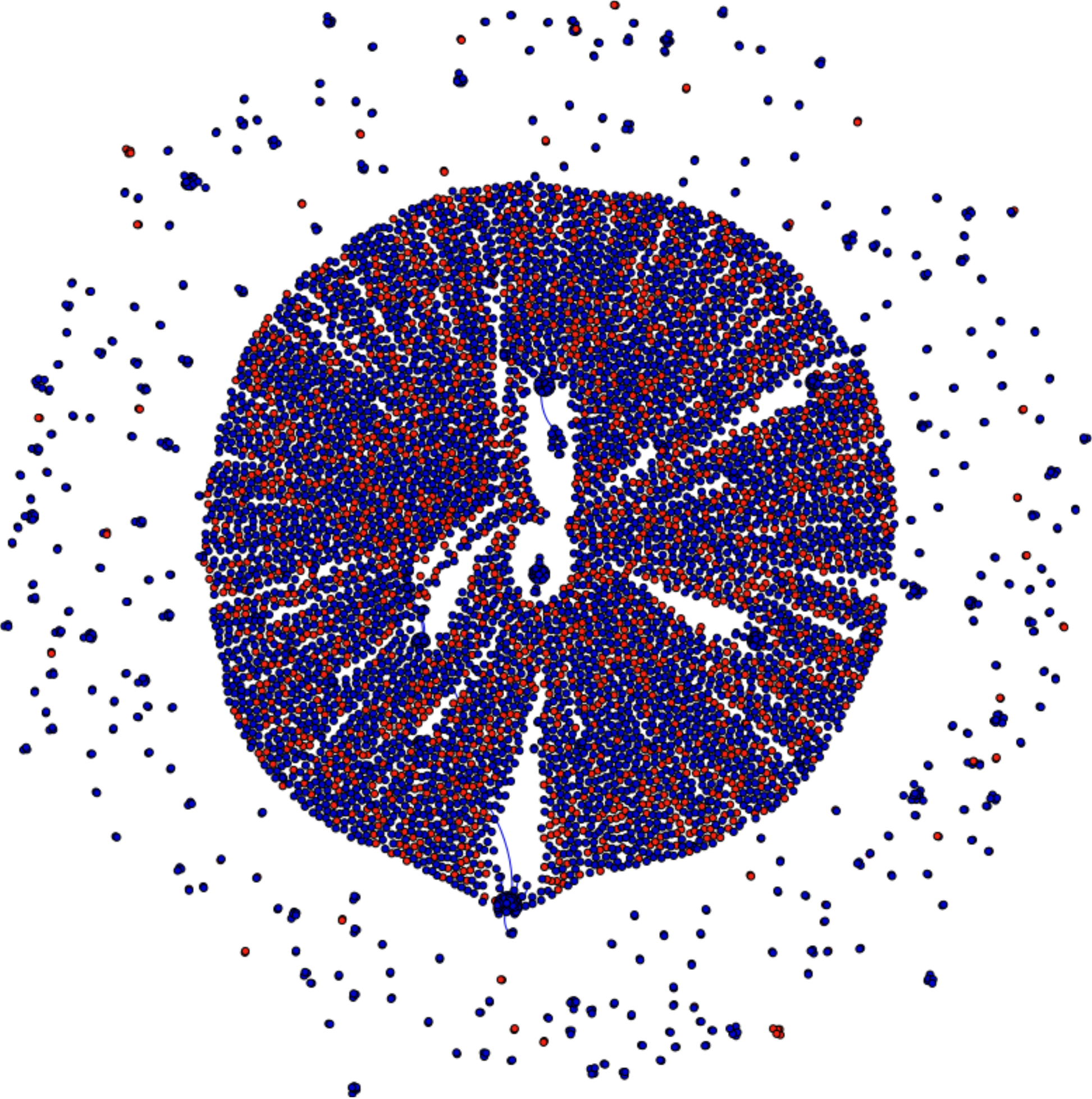}
  \caption{Interactions (Basque Country)}
  \label{fig:interactions-eu}
 \end{subfigure}
 \begin{subfigure}[b]{0.33\textwidth}
  \includegraphics[width=\textwidth]{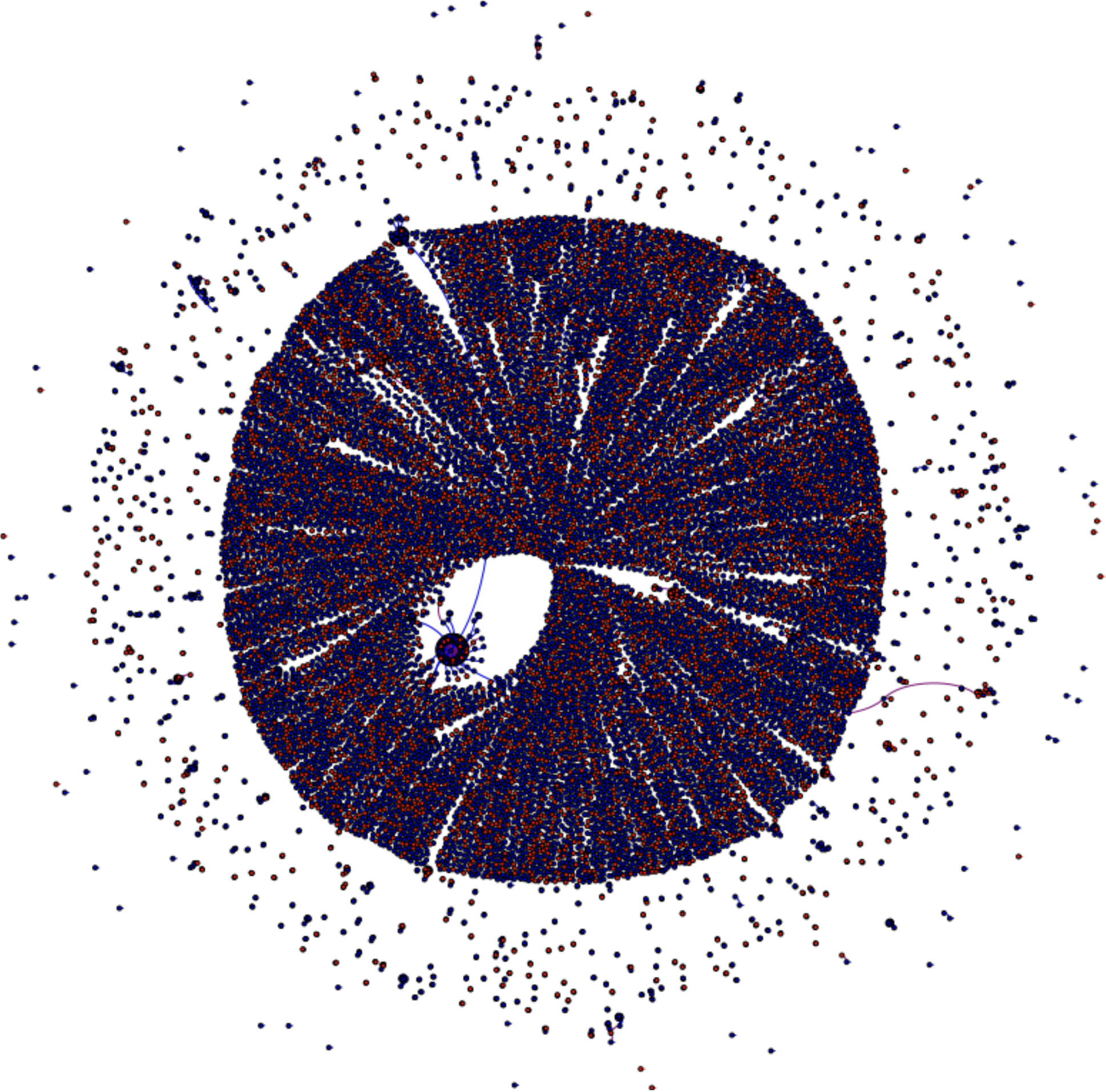}
  \caption{Interactions (Scotland)}
  \label{fig:interactions-sc}
 \end{subfigure}
 \begin{subfigure}[b]{0.33\textwidth}
  \includegraphics[width=\textwidth]{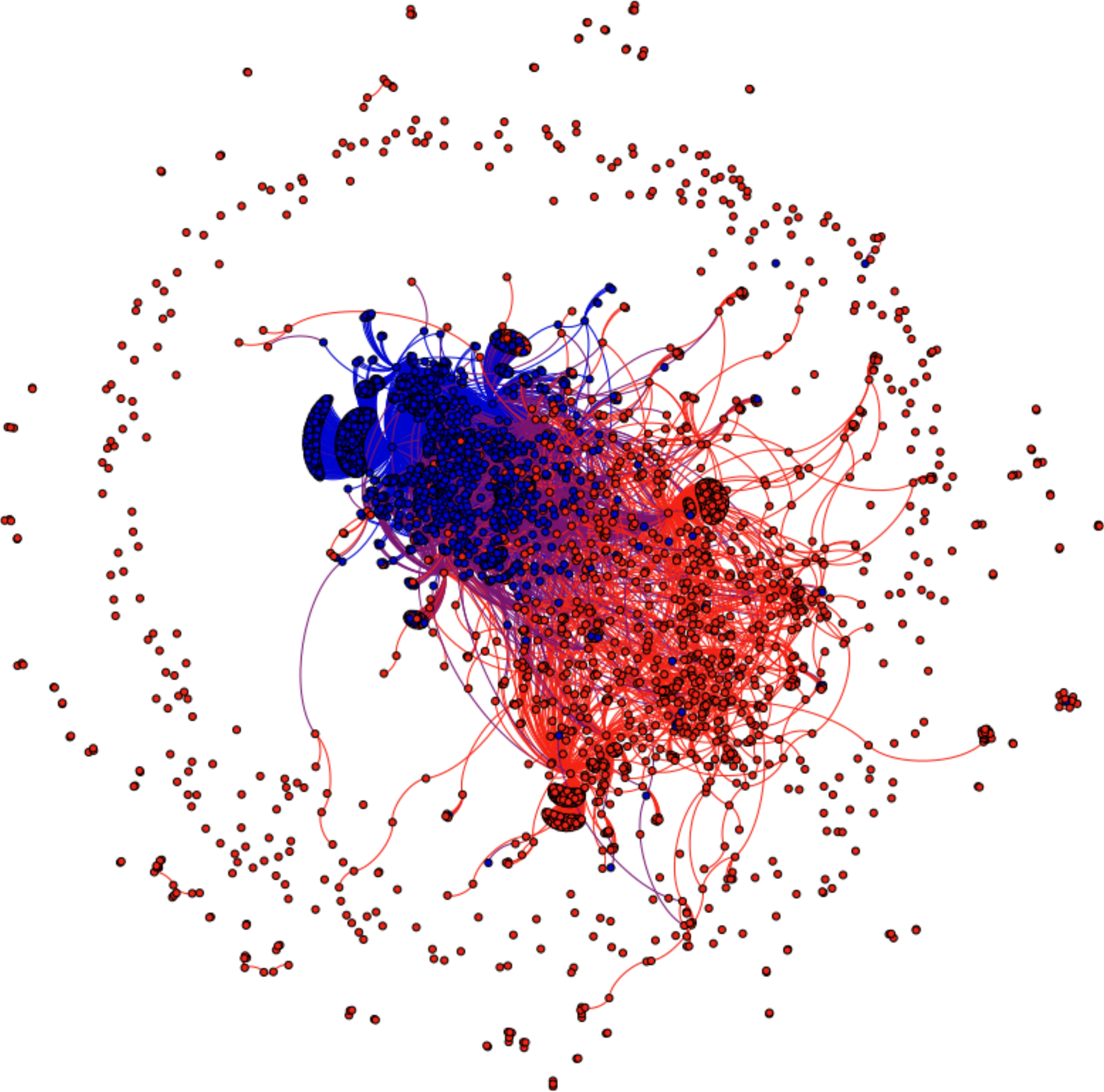}
  \caption{Network (Catalonia)}
  \label{fig:network-ca}
 \end{subfigure}
 \begin{subfigure}[b]{0.33\textwidth}
  \includegraphics[width=\textwidth]{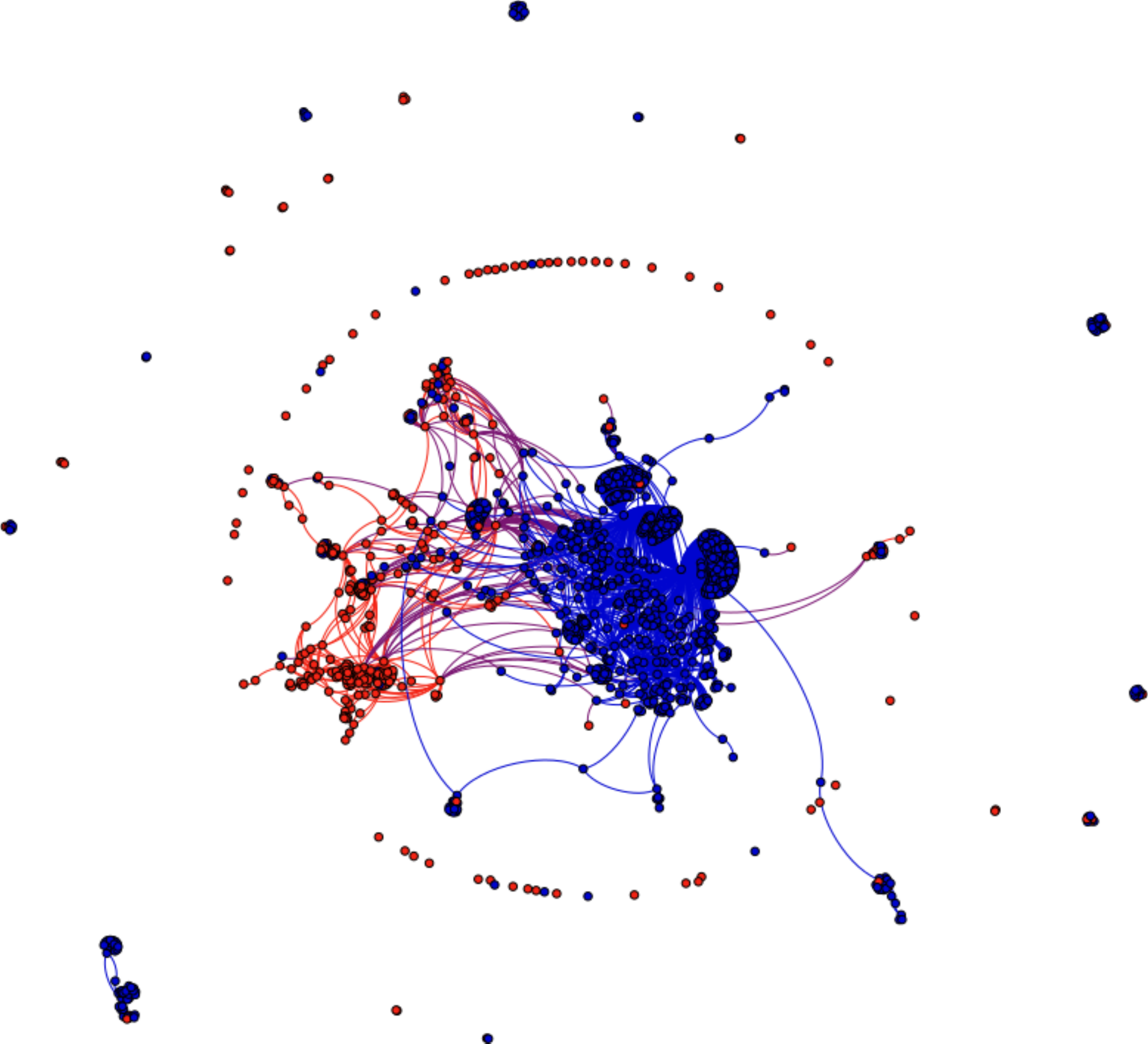}
  \caption{Network (Basque Country)}
  \label{fig:network-eu}
 \end{subfigure}
 \begin{subfigure}[b]{0.33\textwidth}
  \includegraphics[width=\textwidth]{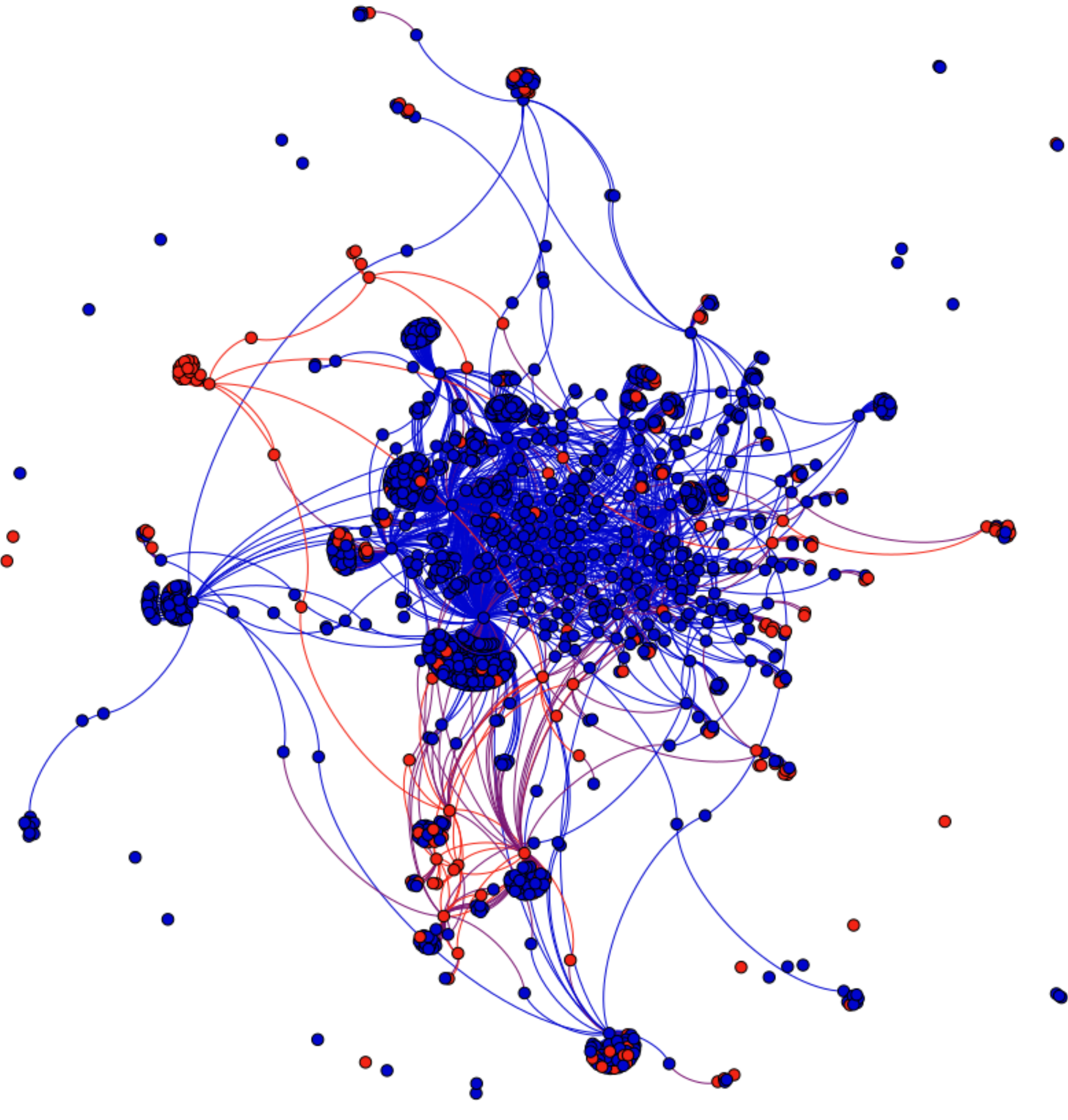}
  \caption{Network (Scotland)}
  \label{fig:network-sc}
 \end{subfigure}
 \caption{Interactions and network connections within and across national identities. Blue: pro-independence; Red: anti-independence.}
 \label{fig:interactions-network}
\end{figure*}

To begin with the analysis of different national identities, in Figure \ref{fig:interactions-network} we look at the interactions and network features by visualising connections within and across national identities. A look at the interactions shows a confusing picture where users of different national identities seem to occasionally interact with each other. However, when we look at the network visualisations, we see a totally different picture where users are mainly connected to others of the same national identity, with a clear separation between national identities, especially for Catalonia and the Basque Country.

To quantify this, we compute the assortativity of these six networks, which is in turn indicative of the existence or not of political homophily \cite{bollen2011happiness}, i.e. the users' preference to connect to and interact with those of the same ideology. Table \ref{tab:assortativity} shows assortativity values for these six networks, along with the analysis of their statistical significance using Mann–Whitney U tests \cite{mann1947test}. All six networks achieve positive assortativity scores indicating statistically significant and positive correlation. These scores are however lower for Scotland, especially in terms of interactions; this suggests that users in Scotland are more likely to follow and interact with each other than in the other two territories, however they still show a preference to follow those who think like them. Separation between communities is much more prominent in the Basque Country and Catalonia, where connections between users who think alike are much more prevalent with assortativity scores above 0.6.

\begin{table}[htb]
 \begin{center}
  \begin{tabular}{l l l}
   \toprule
   & \multicolumn{2}{c}{\textbf{Assortativity (p-value)}} \\
   & \textbf{Network} & \textbf{Interactions} \\
   \midrule
   Catalonia & 0.657 (1.4e-78) & 0.652 (5.0e-161) \\
   \midrule
   Basque Country & 0.678 (5.7e-16) & 0.478 (2.0e-257) \\
   \midrule
   Scotland & 0.311 (1.4e-20) & 0.028 (2.2e-76) \\
   \bottomrule
  \end{tabular}
  \caption{Assortativity or political homophily of follow and interaction networks.}
  \label{tab:assortativity}
 \end{center}
\end{table}

To understand behavioural patterns that characterise the different national identity groups, we perform pairwise comparisons using Welch's t-test \cite{welch1947generalization}. Having two different user groups in each case (pro-independence and anti-independence), Welch's t-test enables us to determine which of the groups is more prominent for a certain feature as well as the statistical significance of that prominence. The results of this analysis are shown in Table \ref{tab:t-tests}, with a set of 30 features grouped into 5 types.

Regarding the tweeting activity of users, we observe that there is no consistent pattern as to who tweets more, has older accounts or gets more retweeted (\#1 to \#6), with pro-independence users being more active in Catalonia, anti-independence users being more active in Scotland and both users being more active in terms of different aspects in the Basque Country. What is interesting is to look at the URLs that these users post in their tweets (\#7 and \#8). We see that pro-independence users tend to post more URLs whose domain belongs to their nation (i.e., .cat for Catalonia, .eus for the Basque Country or .scot for Scotland), whereas anti-independence users tend to post more URLs whose domain belongs to the officially recognised country (.es for Spain and .uk for the UK). This finding is statistically significant for Catalonia and the Basque Country, but not for Scotland.

Looking at the user profiles, we see that pro-independence users tend to have more followers while anti-independence users tend to follow more people in the Basque Country and Scotland, however it is the pro-independence users who have both more followers and follow more people in the case of Catalonia (\#9 and \#10). There is no significant difference when we look at whether users from both groups are verified accounts or not (\#11). The users who have the geolocation feature enabled in their accounts tend to be pro-independence in Catalonia and the Basque Country, and anti-independence in Scotland (\#12). Initially we hypothesised that pro-independence users would be less likely to activate the geolocation feature, given that in that case Twitter would tag their geolocated tweets as coming from Spain or the UK, which they might dislike. However, this only holds true for Scotland and hence the users might not be concerned and/or aware of this.

We also look at the URL specified in the user profiles as being one that belongs to the independent TLD (\#13, .cat/.eus/.scot) or the officially recognised country's TLD (\#14, .es/.uk). We observe significant differences here, for all three territories, showing that pro-independence users tend to use more the independent TLD, with the anti-independence users using more the official country's TLD. Finally, we look at the extent to which the users configure their accounts in the language of the independent nation (\#15) or the official country's language (\#16). There is a significant difference in both Catalonia and the Basque Country, with pro-independence users being more likely to set up their accounts in Catalan and Basque, respectively. This feature is not as indicative for Scotland as Twitter does not allow the option to use the service in Scottish Gaelic or Scots. Instead, our analysis looked at the use of ``en-gb'' as the country's official and ``en'' as the opposite. An analysis with one of the Scottish local languages available in the platform may lead to different results.

Interaction features (\#17-\#24) and network features (\#25-\#27) show a similar tendency; in Catalonia, it is the pro-independence users who are more likely to follow and interact with both groups than the anti-independence users, whereas in the Basque Country and Scotland the pro-independence make more connections within their group and the anti-independence connect more with the opposing group than the pro-independence do. Linguistic features (\#28-\#30) are processed using the Polyglot Python package for language identification and sentiment analysis \cite{chen2014building}. As expected, pro-independence users are more likely to use the language of their territory (Basque, Catalan, Scottish Gaelic or Scots) than the anti-independence, which shows their passion for their cultural background. Looking at the sentiment features (\#29-\#30), however, we do not observe a clear pattern across territories. More interestingly, a comparison of the sentiment in the interactions within and across groups shows that users tweet positively 67.7\% times more often within groups than across groups (MWW = 528932618.0, p $<$ 0.01).

\begin{table*}
 \begin{center}
  \begin{tabular}{c | l | p{2cm} | p{2cm} | p{2cm} }
   \toprule
   Feature \# & \textbf{Feature} & \multicolumn{1}{ c |}{\textbf{Catalonia}} & \multicolumn{1}{ c |}{\textbf{Basque C.}} & \multicolumn{1}{ c }{\textbf{Scotland}} \\
   \hline
   \multicolumn{5}{ c }{\textbf{Tweeting activity}} \\
   \hline
   \#1 & Number of tweets posted & \multicolumn{1}{ c }{\cellcolor{green}\texttt{**}} & \multicolumn{1}{ c }{\cellcolor{green}\texttt{*}} & \multicolumn{1}{ c }{\cellcolor{white}\texttt{**}} \\
   \#2 & Number of tweets favourited & \multicolumn{1}{ c }{\cellcolor{green}\texttt{}} & \multicolumn{1}{ c }{\cellcolor{white}\texttt{**}} & \multicolumn{1}{ c }{\cellcolor{white}\texttt{**}} \\
   \#3 & Tweeting rate (avg. tweets per day) & \multicolumn{1}{ c }{\cellcolor{green}\texttt{**}} & \multicolumn{1}{ c }{\cellcolor{white}\texttt{}} & \multicolumn{1}{ c }{\cellcolor{white}\texttt{**}} \\
   \#4 & Age of the Twitter account & \multicolumn{1}{ c }{\cellcolor{green}\texttt{**}} & \multicolumn{1}{ c }{\cellcolor{green}\texttt{**}} & \multicolumn{1}{ c }{\cellcolor{white}\texttt{**}} \\
   \#5 & Number of retweets their tweets get & \multicolumn{1}{ c }{\cellcolor{green}\texttt{**}} & \multicolumn{1}{ c }{\cellcolor{white}\texttt{**}} & \multicolumn{1}{ c }{\cellcolor{white}\texttt{**}} \\
   \#6 & Number of times their tweets are favourited & \multicolumn{1}{ c }{\cellcolor{green}\texttt{**}} & \multicolumn{1}{ c }{\cellcolor{white}\texttt{**}} & \multicolumn{1}{ c }{\cellcolor{white}\texttt{**}} \\
   \#7 & User posts URLs belonging to independent nation's TLD & \multicolumn{1}{ c }{\cellcolor{green}\texttt{**}} & \multicolumn{1}{ c }{\cellcolor{green}\texttt{**}} & \multicolumn{1}{ c }{\cellcolor{green}\texttt{}} \\
   \#8 & User posts URLs belonging to current country's TLD & \multicolumn{1}{ c }{\cellcolor{white}\texttt{**}} & \multicolumn{1}{ c }{\cellcolor{white}\texttt{**}} & \multicolumn{1}{ c }{\cellcolor{white}\texttt{*}} \\
   
   \hline
   \multicolumn{5}{ c }{\textbf{User profile}} \\
   \hline
   \#9 & Number of accounts that follow them & \multicolumn{1}{ c }{\cellcolor{green}\texttt{**}} & \multicolumn{1}{ c }{\cellcolor{green}\texttt{**}} & \multicolumn{1}{ c }{\cellcolor{green}\texttt{**}} \\
   \#10 & Number of accounts they follow & \multicolumn{1}{ c }{\cellcolor{green}\texttt{**}} & \multicolumn{1}{ c }{\cellcolor{white}\texttt{**}} & \multicolumn{1}{ c }{\cellcolor{white}\texttt{**}} \\
   \#11 & User is verified & \multicolumn{1}{ c }{\cellcolor{white}\texttt{}} & \multicolumn{1}{ c }{\cellcolor{green}\texttt{}} & \multicolumn{1}{ c }{\cellcolor{white}\texttt{}} \\
   \#12 & User has geolocation feature enabled & \multicolumn{1}{ c }{\cellcolor{green}\texttt{**}} & \multicolumn{1}{ c }{\cellcolor{green}\texttt{**}} & \multicolumn{1}{ c }{\cellcolor{white}\texttt{**}} \\
   \#13 & User profile URL belongs to independent nation's TLD & \multicolumn{1}{ c }{\cellcolor{green}\texttt{**}} & \multicolumn{1}{ c }{\cellcolor{green}\texttt{**}} & \multicolumn{1}{ c }{\cellcolor{green}\texttt{**}} \\
   \#14 & User profile URL belongs to current country's TLD & \multicolumn{1}{ c }{\cellcolor{white}\texttt{**}} & \multicolumn{1}{ c }{\cellcolor{white}\texttt{**}} & \multicolumn{1}{ c }{\cellcolor{white}\texttt{**}} \\
   \#15 & User language is that of the independent nation & \multicolumn{1}{ c }{\cellcolor{green}\texttt{**}} & \multicolumn{1}{ c }{\cellcolor{green}\texttt{**}} & \multicolumn{1}{ c }{\cellcolor{green}\texttt{}} \\
   \#16 & User language is that of the current country & \multicolumn{1}{ c }{\cellcolor{white}\texttt{**}} & \multicolumn{1}{ c }{\cellcolor{white}\texttt{**}} & \multicolumn{1}{ c }{\cellcolor{white}\texttt{}} \\
   
   \hline
   \multicolumn{5}{ c }{\textbf{Interactions}} \\
   \hline
   \#17 & Interactions within national identity & \multicolumn{1}{ c }{\cellcolor{green}\texttt{**}} & \multicolumn{1}{ c }{\cellcolor{green}\texttt{**}} & \multicolumn{1}{ c }{\cellcolor{green}\texttt{**}} \\
   \#18 & Interactions across national identities & \multicolumn{1}{ c }{\cellcolor{green}\texttt{**}} & \multicolumn{1}{ c }{\cellcolor{white}\texttt{}} & \multicolumn{1}{ c }{\cellcolor{white}\texttt{**}} \\
   \#19 & Favouriting within national identity & \multicolumn{1}{ c }{\cellcolor{green}\texttt{**}} & \multicolumn{1}{ c }{\cellcolor{green}\texttt{**}} & \multicolumn{1}{ c }{\cellcolor{green}\texttt{**}} \\
   \#20 & Favouriting across national identities & \multicolumn{1}{ c }{\cellcolor{green}\texttt{**}} & \multicolumn{1}{ c }{\cellcolor{white}\texttt{**}} & \multicolumn{1}{ c }{\cellcolor{white}\texttt{**}} \\
   \#21 & Mentions with national identity & \multicolumn{1}{ c }{\cellcolor{green}\texttt{**}} & \multicolumn{1}{ c }{\cellcolor{green}\texttt{**}} & \multicolumn{1}{ c }{\cellcolor{white}\texttt{**}} \\
   \#22 & Mentions across national identities & \multicolumn{1}{ c }{\cellcolor{green}\texttt{**}} & \multicolumn{1}{ c }{\cellcolor{white}\texttt{**}} & \multicolumn{1}{ c }{\cellcolor{white}\texttt{**}} \\
   \#23 & Retweets within national identity & \multicolumn{1}{ c }{\cellcolor{green}\texttt{**}} & \multicolumn{1}{ c }{\cellcolor{green}\texttt{**}} & \multicolumn{1}{ c }{\cellcolor{green}\texttt{**}} \\
   \#24 & Retweets across national identities & \multicolumn{1}{ c }{\cellcolor{green}\texttt{**}} & \multicolumn{1}{ c }{\cellcolor{white}\texttt{}} & \multicolumn{1}{ c }{\cellcolor{white}\texttt{**}} \\
   
   \hline
   \multicolumn{5}{ c }{\textbf{Network}} \\
   \hline
   \#25 & Number of times added to lists by others & \multicolumn{1}{ c }{\cellcolor{green}\texttt{**}} & \multicolumn{1}{ c }{\cellcolor{green}\texttt{}} & \multicolumn{1}{ c }{\cellcolor{green}\texttt{**}} \\
   \#26 & Follow people of their own national identity & \multicolumn{1}{ c }{\cellcolor{green}\texttt{**}} & \multicolumn{1}{ c }{\cellcolor{white}\texttt{**}} & \multicolumn{1}{ c }{\cellcolor{white}\texttt{**}} \\
   \#27 & Follow people of opposing national identity & \multicolumn{1}{ c }{\cellcolor{green}\texttt{**}} & \multicolumn{1}{ c }{\cellcolor{white}\texttt{}} & \multicolumn{1}{ c }{\cellcolor{white}\texttt{**}} \\
   
   \hline
   \multicolumn{5}{ c }{\textbf{Linguistic}} \\
   \hline
   \#28 & User tweets in independent nation's own language & \multicolumn{1}{ c }{\cellcolor{green}\texttt{**}} & \multicolumn{1}{ c }{\cellcolor{green}\texttt{**}} & \multicolumn{1}{ c }{\cellcolor{green}\texttt{**}} \\
   \#29 & Tweets more positively within national identity & \multicolumn{1}{ c }{\cellcolor{white}\texttt{**}} & \multicolumn{1}{ c }{\cellcolor{green}\texttt{**}} & \multicolumn{1}{ c }{\cellcolor{green}\texttt{**}} \\
   \#30 & Tweets more negatively across national identities & \multicolumn{1}{ c }{\cellcolor{white}\texttt{**}} & \multicolumn{1}{ c }{\cellcolor{green}\texttt{**}} & \multicolumn{1}{ c }{\cellcolor{white}\texttt{**}} \\
   
   \bottomrule
  \end{tabular}
 \end{center}
 \caption{Comparison of features across national identities for the three territories under study. Colours indicate the group for which a feature is more prominent: green for more pro-independence, white for more anti-independence. Statistics are computed using Welch's T-tests (** p $<$ 0.01, * p $<$ 0.05).}
 \label{tab:t-tests}
\end{table*}

\section{Stance Classification}

\subsection{Task Definition}
\label{ssec:task-definition}

We formulate the problem of determining the stance of users towards the independence movement in their territory as a binary, supervised classification task. Stance classification of users differs from the increasingly popular stance classification of texts \cite{zubiaga2016coling} in that the stance is explicitly expressed in each text for the latter, while for users one needs to put together behavioural patterns extracted from historical features of their account. The input to the classifier is a set of users from a specific territory. To build the classification model, a training set of users labelled for one of $Y = \{PI, AI\}$ is used ($PI$ = pro-independence, $AI$ = anti-independence). For a test set including a set of new, unseen users, the classifier will have to determine if each of the users is a supporter or opposer of independence, $\hat{Y} = \{PI, AI\}$.

\subsection{Classification Settings}
\label{ssec:classification-settings}

We perform the classification experiments in a stratified, 10-fold cross-validation setting separately for each territory. We micro-average the scores to aggregate the performance across different folds and report the final accuracy scores. We use four different classifiers: Naive Bayes, Support Vector Machines, Random Forests and Maximum Entropy. We use four different types of features, all of which are independent of the location string we used for determining the ground truth:

\begin{compactenum}
 \item \textbf{Timeline:} We use Word2Vec embeddings \cite{mikolov2013distributed} to represent the content of a user's timeline of most recent tweets. The model we use for the embeddings was trained for each territory using the entire collection of tweets. We represent each tweet as the average of the embeddings for each word, and finally get the average of all tweets.
 \item \textbf{Interactions:} We consider that a user is interacting with another when they are retweeting or replying to them. We create a weighted list of all the users that are the target of the interactions in each of our datasets. Given the length of this list, we reduce its size by restricting to the 99th percentile of most common interactions. Each of the remaining users belong to a feature in the resulting vectors. For each user, we represent each of the features in the vectors as the count of interactions the user has had with the user represented by that feature.
 \item \textbf{Favourites:} To represent the content of the tweets favourited by a user, we use the same approach based on word embeddings as for the timeline above, in this case using the content of the tweets favourited by a user instead.
 \item \textbf{Network:} Similar to the approach used for interactions, we aggregate the list of users that appear in the networks (followees or followers) in each of our datasets. We restrict this list to the 99th percentile formed by the most frequent users in each dataset. For each user, we then create a vector with binary values representing whether each of the users is in the network of the current user.
\end{compactenum}

\section{Results}
\label{sec:results}

\begin{table}[htb]
 \begin{center}
  \begin{tabular}{l r r r r}
   \toprule
   & NB & SV & RF & ME \\
   \midrule
   \multicolumn{5}{c}{\textbf{Catalonia}} \\
   \midrule
   Timeline & .940 & .954 & .955 & .944 \\
   Interactions & .841 & .935 & .960 & .946 \\
   Favourites & .919 & .932 & .932 & .923 \\
   Network & .957 & .970 & .965 & \textbf{.972} \\
   \midrule
   \multicolumn{5}{c}{\textbf{Basque Country}} \\
   \midrule
   Timeline & .598 & .867 & .846 & .831 \\
   Interactions & .799 & .826 & .857 & .842 \\
   Favourites & .567 & .819 & .810 & .784 \\
   Network & .889 & .881 & .885 & \textbf{.903} \\
   \midrule
   \multicolumn{5}{c}{\textbf{Scotland}} \\
   \midrule
   Timeline & .595 & .789 & .742 & .724 \\
   Interactions & .620 & .727 & .803 & .779 \\
   Favourites & .546 & .754 & .724 & .720 \\
   Network & .588 & .828 & .830 & \textbf{.849} \\
   \bottomrule
  \end{tabular}
 \end{center}
 \caption{Stance classification results. NB: Naive Bayes, SV: Support Vector Machines, RF: Random Forests, ME: Maximum Entropy.}
 \label{tab:results}
\end{table}

Table \ref{tab:results} shows the classification results. Among the four feature types under study, a user's network is the most indicative feature for determining their stance. This suggests that users belonging to different identity groups tend to be connected to different users on Twitter. The rest of the features are significantly behind the performance of network features, suggesting that the content they engage with and the people they interact with are not as indicative.

Among the classifiers under study, we find that the Maximum Entropy classifier performs better than the rest when network features are used. This is consistent for all three territories, achieving 0.972, 0.903 and 0.849 for Catalonia, the Basque Country and Scotland, respectively.

\section{Discussion}
\label{sec:discussion}

The methodology described here enabled us to gather large datasets to analyse independence movements through social media, developing a classifier that can determine the users' national identity. Our methodology and classifier have been tested in three territories with ongoing independence movements: Scotland, Catalonia and the Basque Country. Our classification experiments show encouraging results with high performance scores that range from 85\% to 97\% in accuracy with the use of a Maximum Entropy classifier that exploits each user's social network. Moreover, an analysis of the social networks of users reveals the existence of political homophily, where users tend to connect with others from the same group or national identity. Further to this experimentation and in a realistic scenario, the classifier trained from users whose self-reported location field reveals their national identity can then be applied to other users in that particular territory. Classification of users by national identity can then be exploited for further analysis of societal and political issues, as well as to target the segment of users according to one's interest.

Our plans for future work include further experimenting our data collection and annotation approach to other territories such as Palestine or Kurdistan.

\section*{Acknowledgments}

This work has been supported by the PHEME FP7 project (grant No. 611233) and The Alan Turing Institute under the EPSRC grant EP/N510129/1.

\bibliographystyle{IEEEtran}
\bibliography{nationalid}

\begin{IEEEbiography}[{\includegraphics[width=1in,height=1.25in,clip,keepaspectratio]{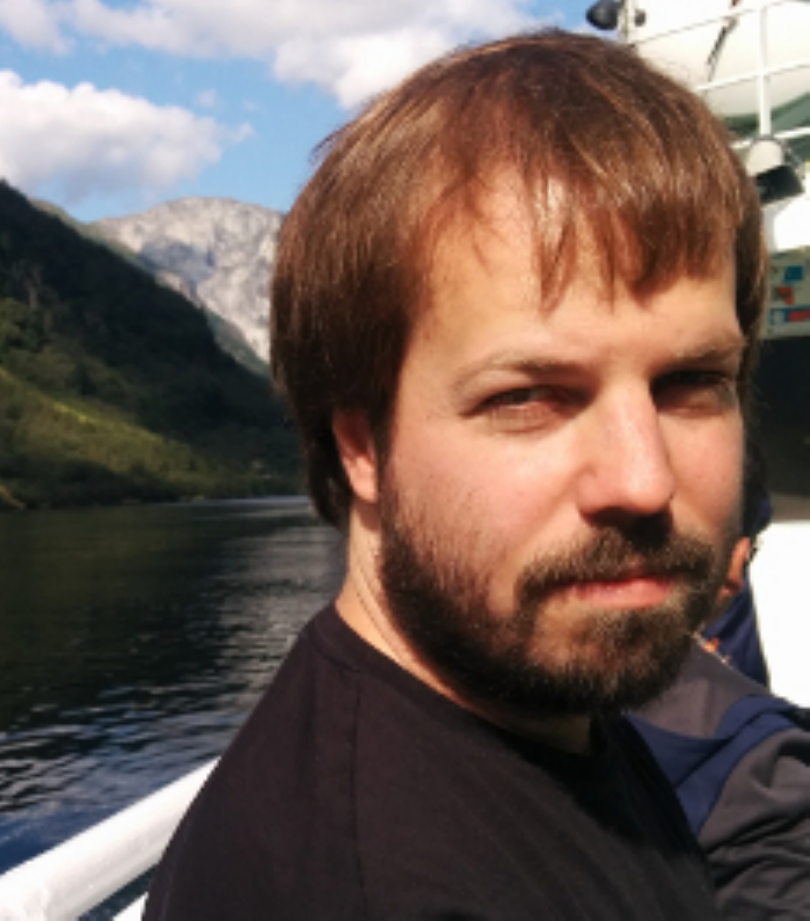}}]{Dr. Arkaitz Zubiaga}
Arkaitz Zubiaga is an assistant professor at the University of Warwick. His research interests revolve around social media mining, natural language processing, computational social science and human-computer interaction.
\end{IEEEbiography}

\begin{IEEEbiography}[{\includegraphics[width=1in,height=1.25in,clip,keepaspectratio]{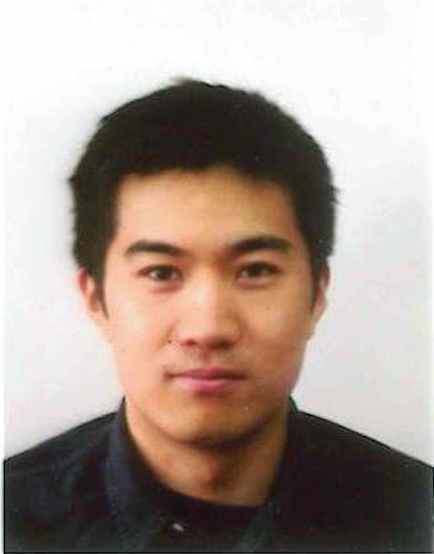}}]{Bo Wang}
Bo Wang is a PhD student at the University of Warwick, under the supervision of Dr. Maria Liakata and Prof. Rob Procter. His research interests lie in social media mining, natural language processing, sentiment analysis and automatic text summarisation. 
\end{IEEEbiography}

\begin{IEEEbiography}[{\includegraphics[width=1in,height=1.25in,clip,keepaspectratio]{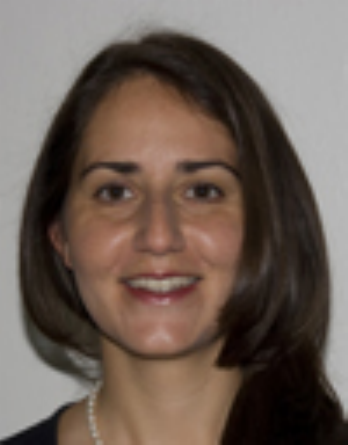}}]{Dr. Maria Liakata}
Maria Liakata is an associate professor at the University of Warwick and a Turing Fellow at the Alan Turing Institute. Her research interests lie in text mining, natural language processing, biomedical text mining and sentiment analysis.
\end{IEEEbiography}

\begin{IEEEbiography}[{\includegraphics[width=1in,height=1.25in,clip,keepaspectratio]{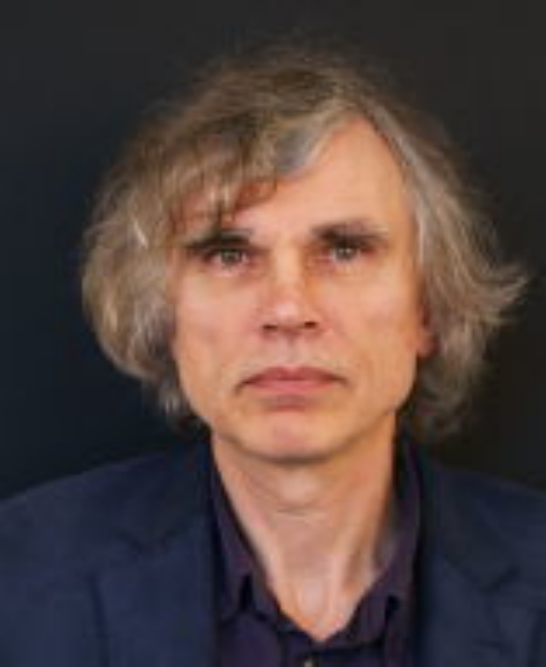}}]{Prof. Rob Procter}
Rob Procter is a Professor at the University of Warwick and a Turing Fellow at the Alan Turing Institute. His research interests lie in social media analysis, computational social science, natural language processing and mixed methods.
\end{IEEEbiography}

\end{document}